\documentclass[10pt]{article} 
\usepackage[preprint]{tmlr}

\usepackage{amsmath,amsfonts,bm}

\def\eqref#1{equation~\ref{#1}}

\def\1{\bm{1}}

\DeclareMathAlphabet{\mathsfit}{\encodingdefault}{\sfdefault}{m}{sl}
\SetMathAlphabet{\mathsfit}{bold}{\encodingdefault}{\sfdefault}{bx}{n}

\def\sI{{\mathbb{I}}}

\usepackage{hyperref}
\usepackage{url}
\usepackage{graphicx}

\usepackage{breakcites}
\makeatletter
\newcommand{\printfnsymbol}[1]{%
\textsuperscript{\@fnsymbol{#1}}%
}

\title{Decentralized Autoregressive Generation}

\author{\name Stepan Maschan \email s.maschan@lancaster.ac.uk \\
      \addr School of Computing and Communications\\
      Lancaster University \AND \name Haoxuan Qu \thanks{Corresponding authors} \email h.qu5@lancaster.ac.uk \\
      \addr School of Computing and Communications\\Lancaster University \AND \name Jun Liu \printfnsymbol{1} \email j.liu81@lancaster.ac.uk \\\addr School of Computing and Communications\\Lancaster University}

\newtheorem{thm}{Theorem}
\newcommand{\m}{\mathbf{m}}

\def\month{MM}  
\def\year{YYYY} 
\def\openreview{\url{https://openreview.net/forum?id=XXXX}} 

\begin{document}

\maketitle

\begin{abstract}
The decentralization of autoregressive generation has attracted considerable attention in recent years as a solution to scaling bottlenecks. However, despite promising empirical results, this paradigm currently lacks rigorous theoretical justification. In this work, we formally establish the theoretical equivalence between decentralized and centralized training. To achieve this, we adapt the Discrete Flow Matching framework for autoregressive generation, leveraging its inherent properties to demonstrate that global models naturally decompose into independent experts. Finally, we conduct extensive experiments across diverse multimodal benchmarks, empirically validating that decentralized training maintains competitive parity with standard centralized architectures.
\end{abstract}

\section{Introduction}

Autoregressive generation has established itself as the dominant paradigm for a diverse range of machine learning tasks, encompassing natural language modeling, multimodal reasoning, robotics, and synthetic media generation \citep{comanici2025gemini25pushingfrontier, brohan2023rt2visionlanguageactionmodelstransfer, zhou2025indextts2breakthroughemotionallyexpressive}. The efficacy of these models is characterized by scaling laws \citep{henighan2020scalinglawsautoregressivegenerative, kaplan2020scalinglawsneurallanguage}, which demonstrate that increasing the parameter count of a model and the amount of training data predictably improves its performance.

However, this necessary increase in parameter counts leads to significant infrastructural challenges. The standard approach relies on centralized training, which involves training a single large model across a vast cluster of GPU nodes. To function effectively, these thousands of GPUs have to continuously communicate with one another to synchronize gradients. This massive hardware requirement creates a severe accessibility barrier, making it extremely difficult for academic institutions and smaller organizations to participate in foundational model training. Furthermore, this tightly coupled setup is highly vulnerable to individual GPU node failures. In a centralized system, if even a single GPU fails, it disrupts the global training process; as noted by \citet{grattafiori2024llama3herdmodels}, this often requires a restart of the entire training job from the last saved checkpoint, wasting significant time and computational resources.

A promising solution to these hardware bottlenecks is decentralized training. Instead of training one model on all the data, this approach trains a collection of independent "expert" models. The process works in two stages. First, the massive global dataset is divided into separate, non-overlapping subsets. Each expert model is then trained exclusively on one of these subsets. Because the experts never need to communicate during training, they can be trained on completely separate, smaller computer systems. Second, during inference, a router evaluates the input and activates only one or a few the most relevant experts. While this router is often trained as a separate auxiliary model, it can also be derived directly from the mechanisms used to partition the data. Their outputs are then combined to generate the final response. This approach removes the need for a vast centralized computation and data clusters and makes collaborative training feasible for researchers with limited hardware.

Recent studies have shown that decentralized training is highly effective for text generation models in certain situations \citep{gururangan2023scalingexpertlanguagemodels, no-need-to-talk, li2022branchtrainmergeembarrassinglyparalleltraining}. However, these successes are largely based on practical experiments rather than strict theoretical foundations. Presently, it remains unknown whether the decentralized and centralized training paradigms are mathematically equivalent. Consequently, decentralized training is often treated as a heuristic "black box". Without a theoretical foundation researchers lack any a priori expectations of model convergence or capability. They can only assess the success of a decentralized approach retrospectively, after the training is complete. This theoretical uncertainty highlights a significant gap in the literature and motivates a critical question: can we theoretically guarantee that a decentralized collection of experts is equivalent to a standard, centrally trained model?

To answer this question, we seek to establish a rigorous theoretical foundation for decentralized training. During exploration, we find a critical insight from recent advancements in continuous generative modeling: within the standard Flow Matching framework, the centralized and decentralized generating flows have already been shown to be mathematically equivalent \citep{mcallister2025decentralizeddiffusionmodels}. This established equivalence presents a promising pathway. If autoregressive generation can be reformulated through the lens of Flow Matching, it could naturally inherit these same theoretical guarantees. To achieve this, we turn to the theory of Discrete Flow Matching \citep{campbell2024generativeflowsdiscretestatespaces, gat2024discrete}. While adapting the Discrete Flow Matching framework to the discrete-time mechanics of autoregressive generation is non-trivial, it enables us to express the generative process using the "probability generating velocity", the discrete analogue of a continuous generative flow. We demonstrate that this global velocity can be structurally decomposed and represented as a strict linear combination of local velocities produced by independent data clusters. By proving this mathematical decomposition, we move beyond empirical heuristics and formally establish the theoretical equivalence of the centralized and decentralized generating objectives for autoregressive models.

To realize this reformulation, it is necessary and sufficient to show that autoregressive generation is a special case of Discrete Flow Matching. However, establishing this link requires addressing a fundamental difference in how they operate. Standard flow matching assumes that probability distributions change smoothly over continuous time, whereas autoregressive generation relies on discrete, step-wise changes. Because of these non-smooth transitions, the core mathematical guarantee of flow matching, specifically, the link between the Continuity Equation and the generated probability path does not hold in an arbitrary discrete-time case. To resolve this, we demonstrate that imposing a 1-sparse constraint on the probability generating velocity (i.e., restricting probability changes to a single token position at any given timestep) restores the theoretical validity of the framework. By explicitly enforcing this property, we prove that autoregressive generation perfectly satisfies the Discrete Flow Matching paradigm. We provide a comprehensive theoretical analysis of this formulation in Section 4.

Our primary contributions are as follows:
\begin{itemize}
\item For the first time, we establish a rigorous theoretical equivalence between decentralized and centralized training paradigms for autoregressive generation. By proving that the global probability generating velocity can be exactly expressed as a linear combination of independent expert velocities, we provide the theoretical analysis that justifies decentralized training.
\item To achieve this, we extend the Discrete Flow Matching framework into the discrete-time domain and formally align it with autoregressive generation. We introduce the necessary constraints, such as the 1-sparse property, to make this alignment possible. Beyond enabling our primary proof, this formulation serves as a novel, standalone mathematical bridge between continuous flow-based methods and discrete autoregressive models.
\item We provide extensive empirical evidence demonstrating that this decentralized paradigm remains highly effective beyond standard text generation. We successfully apply decentralized training to Multimodal Large Language Models (MLLMs), conducting extensive experiments across multiple question-answering benchmarks utilizing the LLaVA 
\citep{liu2024improvedbaselinesvisualinstruction} and InternVL \citep{chen2025expandingperformanceboundariesopensource} architectures.
\end{itemize}
\section{Related Work}
\paragraph{Autoregressive Generation} Autoregressive generation has become a state-of-the-art paradigm for multiple ML tasks, such as language modeling, multimodal reasoning, robotics, image generation, audio synthesis and others \citep{vaswani2023attentionneed, grattafiori2024llama3herdmodels, comanici2025gemini25pushingfrontier, henighan2020scalinglawsautoregressivegenerative, kaplan2020scalinglawsneurallanguage, Guo_2025, brohan2023rt2visionlanguageactionmodelstransfer, tian2024visual, zhou2025indextts2breakthroughemotionallyexpressive, liu2024improvedbaselinesvisualinstruction, chen2025expandingperformanceboundariesopensource}. For example, \citet{brohan2023rt2visionlanguageactionmodelstransfer} finetuned a pretrained visual language model on a mix of original training data and robot data, where robot actions were represented as tokens, resulting in significant generalization improvements. Another work, \citet{tian2024visual}, used an autoregressive approach to the image generation, using image scales of increasing resolutions as autoregressive units. Each next scale is predicted using only its prefix of previous scales. This approach provided a superior image generation performance to the previous work. Besides that, autoregressive generation showed its efficiency in visual question answering. As shown in \citet{liu2024improvedbaselinesvisualinstruction}, finetuning pretrained language model on a relatively small VQA dataset, where image features are projected into token space through Multilayer Perceptron, can lead to SOTA results on various VQA tasks. \citet{chen2025expandingperformanceboundariesopensource} further demonstrated that this approach is capable of delivering a performance comparable to native multimodal models. The scaling laws \citep{kaplan2020scalinglawsneurallanguage, henighan2020scalinglawsautoregressivegenerative} allow us to predict model performance increase through the growth of model parameters and training dataset size. However, the maintenance of large amounts of data and training large centralized models across multiple GPUs presents significant infrastructural challenges \citep{grattafiori2024llama3herdmodels}. Because standard centralized training requires highly synchronous communication, the system is vulnerable to GPU node failures, which can halt the entire training process. Furthermore, this paradigm imposes severe engineering demands for storing and managing colossal, centralized datasets across high-bandwidth networks.
\paragraph{Decentralized  Training}
Decentralized training has emerged in the recent studies as an effective way to drastically reduce communication overhead between training nodes \citep{gururangan2023scalingexpertlanguagemodels, no-need-to-talk, ma2024modeclipdataexperts, mcallister2025decentralizeddiffusionmodels, wang2025federatedflowmatching, li2022branchtrainmergeembarrassinglyparalleltraining}. For example, \citet{ma2024modeclipdataexperts} trained a mixture of CLIP model experts independently on different data clusters, reducing the amount of false negative examples in each cluster. Another notable example is Decentralized Diffusion Models framework \citep{mcallister2025decentralizeddiffusionmodels}, which uses a mixture of flow matching expert models for image generation to outperform dense baseline model. Another approach to decentralized flow matching is Federated Flow Matching \citep{wang2025federatedflowmatching}, using semi-dual formulation of optimal transport to approximate global optimal coupling, leading to straighter probability paths and thereby superior performance compared to vanilla federated gradient averaging. While previous work empirically verifies the effectiveness of decentralized training and inference in certain scenarios, our work for the first time suggests a theoretical analysis of the decentralized autoregressive generation in unified settings. In addition, prior work focuses either on pure language modeling, image
generation or contrastive learning, while our work conducts experiments for multimodal large language models for visual question answering.

\paragraph{Discrete Flow Matching}
Discrete Flow Matching extends Flow Matching theory \citep{lipman2023flow} to discrete probability spaces. It gained attention in text modeling as an alternative to autoregressive generation, protein co-design, graph generation and others \citep{campbell2024generativeflowsdiscretestatespaces, qin2025defog, gat2024discrete}. Notably, \citet{gat2024discrete} considered probability paths that are convex sums of elementary probabilities and provided the closed formula for probability generating velocity. \citet{davis2024fisherflowmatchinggenerative} introduced an alternative theoretical foundation, by mapping probability manifold with Fisher-Rao metric to a sphere and considering probability paths as geodesics on sphere. While prior work provides substantial background for continuous time probability paths, the discrete time probability paths case is not investigated. In this work, we extend Discrete Flow Matching to the discrete time case, as required for autoregressive scenario, reformulate the equations and investigate the constraints under which probability generating velocities for discrete time paths are correctly defined.

\section{Background}
As discussed, our central objective is to establish a rigorous theoretical foundation for decentralized autoregressive generation. We begin by observing that autoregressive generation is, at its core, a specific instance of a probability path moving between distributions in a discrete token space. Therefore, mathematically formalizing and ultimately decentralizing this process requires a framework explicitly designed for discrete probability dynamics. This section briefly reviews the standard Discrete Flow Matching framework \citep{gat2024discrete}. We specifically focus on defining probability paths, probability generating velocities, and the Continuity Equation. We highlight these exact concepts because the probability path formalizes the sequence's trajectory from source to target, the generating velocity dictates the transition rules at each step, and the Continuity Equation provides the mathematical guarantee that following this velocity yields the correct target distribution. Formally reintroducing these standard properties is an essential prerequisite before we can adapt them to the discrete-time mechanics of autoregressive models in our subsequent theoretical analysis.

\subsection{Discrete Flow Matching (Continuous Time)}
Discrete Flow Matching \citep{gat2024discrete} is a theoretical framework that describes discrete token generation processes in terms similar to Continuous Flow Matching \citep{lipman2023flow}. In this subsection we introduce foundational concepts from this framework that are necessary for our theoretical analysis in Section \ref{sec:4}. In the following paragraphs, we will define the probability path, the probability generating velocity, and the Continuity Equation.

\paragraph{Probability paths} Given vocabulary of tokens $[d]$, source distribution $X_0$ with probability mass function $p(x)$ of token sequences $x=(x^1, \ldots, x^N)\in[d]^N$ and target distribution $X_1$ with probability mass function $q(x)$, we define the \textbf{probability path} $p_t(x), t\in[0,1]$ s.t. $p_0(x)=p(x)$, $p_1(x)=q(x)$ as follows:
\begin{gather}\label{eq:1}
    p_t(x)=\sum_{(x_0, x_1)\in[d]^{N}\times [d]^{N}}p_t(x|x_0, x_1)\pi(x_0, x_1) \\
    \label{eq:2}
    p_t(x|x_0, x_1)=\prod_{i=1,\ldots ,N}p_t(x^i|x_0, x_1)
\end{gather}
where $\pi(x_0, x_1)$ is a joint probability mass function of $(X_0, X_1)$, and $p_t(x^i|x_0, x_1)$ is a conditional marginal probability path for the token position $i$. 

The $p_t(x|x_0,x_1)$ should satisfy the following conditions:
\begin{gather}
    p_0(x|x_0, x_1)=\delta_{x_0}(x)=\begin{cases}
        1, & x=x_0 \\
        0, & x\neq x_0
    \end{cases} \\
    p_1(x|x_0, x_1)=\delta_{x_1}(x)=\begin{cases}
        1, & x=x_1 \\
        0, & x\neq x_1
    \end{cases} 
\end{gather}
In the framework, in order for probability paths to be tractable, we consider only the probability paths that are represented by a convex sum of conditional probabilities $w^j(x^i|x_0, x_1)$ with coefficients provided by some scheduler $\kappa^{i,j}_t$:
\begin{gather}\label{eq:5}
    p_t(x^i|x_0, x_1)=\sum_j \kappa^{i,j}_t w^j(x^i|x_0, x_1) \\
    \label{eq:6}
    \sum_j\kappa^{i, j}_t=1,\quad \kappa_t^{i, j}\geq 0
\end{gather}

\paragraph{Probability generating velocities}
We also define \textbf{probability generating velocity} $u^i_t(x,z),i\in[N]$. We sample our token sequences by the following rule (for each token position independently):
\begin{gather}\label{eq:7}
    X_{t+h}^i\sim\delta_{X_t^i}(\cdot)+hu_t^i(\cdot,X_t),\quad h>0
\end{gather}
We say that $u_t$ generates the probability path $p_t(x)$ if $X_{t+h}$, sampled by the rule \ref{eq:7}, satisfies:
\begin{gather}\label{eq:8}
    X_{t+h}\sim p_{t+h}+o(h)
\end{gather}
Probability generating velocity $u_t$ can be constructed through conditional probability generating velocities $u^i_t(x|x_0,x_1)$. In fact, the following theorem holds:
\begin{thm}\label{theorem1}
    Probability generating velocity $u_t^i(x^i, z)$ defined as follows:
     \begin{gather}\label{eq:9}
         u_t^i(x^i, z)=\sum_{(x_0, x_1)\in [d]^{N}\times [d]^{N}}u_t^i(x^i,z|x_0,x_1)\frac{p_t(z|x_0,x_1)\pi(x_0,x_1)}{p_t(z)}
     \end{gather}
    generates the probability path $p_t(x)$.
\end{thm}

\paragraph{Continuity Equation} Theorem \ref{theorem1} is proven by using the Continuity Equation:
\begin{gather}
    \dot{p_t}(x)+div_x(p_tu_t)=0 
\end{gather}
The discrete divergence of $p_tu_t$ is defined in the following way:
\begin{gather}
    div_x(v)=\sum_{z\in [d]^{N}} (v(z, x)-v(x,z))
\end{gather}
where   $v(z,x)=p_t(x)u_t^i(z^i, x)$, $v(x,z)=p_t(z)u_t^i(x^i,z)$ if $z$ differs from $x$ only in the token position $i$, $v(x,x)=p_t(x)\sum_{i=1}^Nu_t^i(x^i,x)$ and $v(z,x)=v(x,z)=0$ otherwise.

The divergence of $p_t u_t$ can be shown to be equal to:
\begin{gather}\label{eq:12}
    div_x(p_tu_t)=-\sum_{z\in[d]^N}p_t(z)\sum_{i=1}^N\delta_{z}(x^{\bar{i}})u_t^i(x^i, z)
\end{gather}
where $\delta_z(x^{\bar{i}})=\prod_{j\neq i}\delta_z(x^j)$.

As shown in \citet{gat2024discrete}, if the continuity equation holds, then the probability generating velocity generates a probability path as defined in (\ref{eq:7}-\ref{eq:8}).
\section{Theoretical Analysis}\label{sec:4}
The primary objective of this section is to mathematically prove that the training objective for autoregressive sampling can be completely decentralized. To achieve this, we formulate autoregressive generation as a special case of Discrete Flow Matching. By establishing this theoretical connection, we can demonstrate that the Discrete Flow Matching objective is inherently decentralizable, which subsequently allows us to apply a decentralized formulation directly to autoregressive generation. To build this mathematical proof, we proceed as follows. First, in Subsection \ref{subsec:dfmdt}, because autoregressive generation operates in discrete time steps, we extend the Discrete Flow Matching framework into the discrete time domain. Next, in Subsection \ref{subsec:ars}, we formalize the probability path and the probability generating velocity specifically for autoregressive sampling. Since verifying that the probability generating velocity correctly yields the probability path reduces to checking the Continuity Equation, we establish a necessary and sufficient 1-sparse condition that the discrete velocity must satisfy. Finally, in Subsection \ref{subsec:decent}, we define the Discrete Decentralized Flow Matching objective, which naturally applies to autoregressive generation, thereby completing our proof of theoretical equivalence between the decentralized and centralized training settings.

\subsection{Discrete Flow Matching (Discrete Time)}\label{subsec:dfmdt}
In this subsection, we formally extend the Discrete Flow Matching framework into the discrete time domain. We sequentially introduce the discrete sampling rule, the necessary conditions for the probability generating velocity, and the adapted Continuity Equation. Once these foundational elements are established, we will be able to specifically define the probability paths and the probability generating velocities, and ultimately confirm that the velocity correctly generates the path by verifying the Continuity Equation.

We begin by transitioning to a discrete-time domain, $t \in \{0,1,\ldots,n\}$, and correspondingly redefining the terms. In this setting, the sampling rule~\ref{eq:7} becomes:
\begin{gather}\label{eq:13}
    X_{t+1}^i\sim\delta_{X_t^i}(\cdot)+u_t^i(\cdot, X_t),\quad t\in \{0,1,\ldots,n-1\}
\end{gather}
where probability generating velocity $u_t$ is now defined in discrete timesteps. $u_t$ now generates $p_t(x)$ if $X_{t+1}$ sampled by the rule \ref{eq:13} satisfies the following:
\begin{gather}\label{eq:14}
    X_{t+1}\sim p_{t+1}
\end{gather}
Here, in order to define proper probability mass function, it is necessary and sufficient that probability velocity $u_t$ satisfies the following conditions:
\begin{gather}
    \sum_{x^i\in[d]}u_t^i(x^i,z)=0 \\
    -1 \leq u_t^i(z^i,z)\leq 0,\quad 0\leq u_t^i(x^i,z)\leq 1\quad \text{for} \;x^i\neq z^i
\end{gather}
The continuity equation now takes the form: 
\begin{gather}
    p_{t+1}(x)-p_t(x)+div_x(p_tu_t)=0
\end{gather}

Consequently, confirming that the probability generating velocity correctly yields the probability path reduces entirely to verifying this Continuity Equation. However, an arbitrary velocity will not inherently satisfy this requirement; to guarantee a valid generation process, the velocity must meet a specific 1-sparse condition, which we detail in Subsection \ref{subsec:ars}.

\subsection{Autoregressive sampling}\label{subsec:ars}
In this subsection, we formulate autoregressive sampling as a special instance of the discrete time Discrete Flow Matching framework established in Subsection \ref{subsec:dfmdt}. Following this theoretical foundation, we sequentially introduce the source and target distributions, the conditional probability paths, and the corresponding schedulers. Next, we establish the conditional probability generating velocity and verify that the Continuity Equation holds under these conditions. Finally, we demonstrate that when the probability generating velocity satisfies the 1-sparse property, the Continuity Equation becomes strictly equivalent to the generating probability path. Through this sequence of proofs, we formally validate that autoregressive sampling is a correct instance of Discrete Flow Matching.

To define autoregressive sampling, we define the source and target distribution as a special case of C-coupling from \citet{gat2024discrete}: 
\begin{gather}
    (X_0, X_1)\sim(\sI \odot X_1+(1-\sI)\odot (\m, \m, \ldots, \m), X_1)
\end{gather}
where $X_1\sim q$, $\m$ is a mask token, $\sI\in\{0,1\}^N$ is an indicator vector of the form $\sI=(1, 1, \dots, 1, 0, \dots, 0)$ which has all ones before some token position and all zeros after. 

Next, the conditional probability path for a pair $x_0, x_1$ s.t. $x_0$ has $P$ tokens revealed at timestep 0 will take the form:
\begin{gather}\label{eq:17}
    p_t(x^i | x_0, x_1)=\kappa_t^i\delta_{x_1}(x^i)+(1-\kappa_t^i)\delta_{x_0}(x^i) \\
    \kappa_t^i=\begin{cases}
        0, & t < i - P\\
        1, & t \geq i - P
    \end{cases}
\end{gather}
This definition of conditional probability path and scheduler simply means that at timestep $t$ exactly $t+P$ tokens are revealed.

Note that at any time step $t$, $p_t(x|x_0, x_1)$ is a degenerate distribution with a single result $x_t$: \begin{gather}
    p_t(x|x_0, x_1)=\delta_{x_t}(x)
\end{gather}

Now, define the conditional generating velocity $u_t^i(x^i, z|x_0, x_1)$ as:
\begin{gather}\label{eq:19}
    u_t^i(x^i, z|x_0, x_1)=\begin{cases}
        \delta_{x_{t+1}}(x^i)-\delta_{x_t}(x^i), & z=x_t \\
        0, & z\neq x_t
    \end{cases}
\end{gather}
From the above formula it follows that for a fixed timestep $t$, $u_t(x^i,x_t|x_0,x_1)= 0$ for any token position except possibly $i=P+t+1$. This property is necessary and sufficient to show that the continuity equation implies the generation of probability path by probability generation velocity in the discrete time domain.

Next, in order to show that defined conditional probability generating velocity generates conditional probability path, we need to 1) Show that continuity equation holds 2) Show that if the continuity equation holds, then generation follows.

First, check that the continuity equation holds:
\begin{gather}
    p_{t+1}(x|x_0,x_1)-p_t(x|x_0, x_1)+div_x(p_tu_t)=0
\end{gather}
Since $p_t(x|x_0,x_1)$ has only one outcome $x_t$ and $u_t^i=0$ for any $i\neq P+t+1$, the divergence formula \ref{eq:12} simplifies to the following:
\begin{gather}
    div_x(p_tu_t)=-\delta_{x_t}(x^{\overline{P+t+1}})u_t(x^{P+t+1},x_t|x_0,x_1)
\end{gather}
Substituting \ref{eq:19}, we get:
\begin{align*}
    &p_{t+1}(x|x_0,x_1)-p_t(x|x_0, x_1)+div_x(p_tu_t)\\
    &=
    \delta_{x_{t+1}}(x)-\delta_{x_t}(x)-\delta_{x_t}(x^{\overline{P+t+1}})u_t(x^{P+t+1},x_t|x_0,x_1)\\
    &=
    \delta_{x_{t+1}}(x)-\delta_{x_t}(x)-\delta_{x_t}(x^{\overline{P+t+1}})(\delta_{x_{t+1}}(x^{P+t+1})-\delta_{x_t}(x^{P+t+1}))\\
    &=
    \delta_{x_{t+1}}(x)-\delta_{x_{t}}(x)-\delta_{x_{t+1}}(x)+\delta_{x_{t}}(x)\\
    &=0
\end{align*}
Now, once we verified that the continuity equation holds, we show that satisfying the continuity equation implies generation. For that, we require a 1-sparsity property of generating probability velocity: for a fixed timestep $t$ $u_t^i=0$ for any token position $i$ except possibly one, denoted by $j=j(t)$ depending only on a timestep $t$. As discussed above, velocity defined in (\ref{eq:19}) satisfies it. Then the PMF of $X_{t+1}$ sampled by \ref{eq:13} is: 
\begin{align*}
    p_{X_{t+1}}(x)=& \sum_{z\in[d]^N} p_t(z)\prod_{i=1}^N\bigl(\delta_z(x^i)+u_t^i(x^i, z)\bigr)\\
    &= \sum_{z\in[d]^N} p_t(z)\Bigl(\delta_z(x)+\delta_z(x^{\bar{j}})\,u_t^j(x^j,z)\Bigr)\\
    &= p_t(x)+\sum_{z\in[d]^N} p_t(z)\,\delta_z(x^{\bar{j}})\,u_t^j(x^j,z)\\
    &= p_t(x)-\operatorname{div}_x\bigl(p_t u_t\bigr)=p_t(x)-\left(p_{t}(x)-p_{t+1}(x)\right)=p_{t+1}(x)
\end{align*}

We defined autoregressive sampling as an instance of discrete flow matching in a discrete time domain.  We proved that for probability generating velocities that are non-zero at only one token position for any fixed timestep satisfying continuity equation implies generation. We showed that the autoregressive probability path defined in \ref{eq:17} and the velocity defined in \ref{eq:19} satisfy the continuity equation and therefore define a valid probability path and generating velocity. 
\subsection{Decentralization}\label{subsec:decent}
In this subsection, we demonstrate that the Discrete Flow Matching training objective can be decentralized, and, by extension, that autoregressive sampling is also decentralizable. Specifically, we formally represent the global probability generating velocity as a linear combination of independent expert velocities. Because autoregressive sampling was proven to be an instance of Discrete Flow Matching in Subsection \ref{subsec:ars}, this mathematical decomposition automatically establishes that the autoregressive sampling objective is fully decentralizable. This finalizes our theoretical framework, completing our overarching proof that the decentralized training objective is mathematically equivalent to the centralized baseline.

The formalization from Subsection \ref{subsec:ars} automatically makes Theorem \ref{theorem1} true for autoregressive sampling. Now, similar to \citet{mcallister2025decentralizeddiffusionmodels}, if we define a target distribution $X_1$ as a union of disjoint clusters $S_k,\quad k\in\{1,2,\dots,K\}$, we can rewrite Equation \ref{eq:9} as follows:
     \begin{align}\label{eq:25}
         & u_t^i(x^i, z)=\sum_{(x_0, x_1)\in [d]^{N}\times [d]^{N}}u_t^i(x^i,z|x_0,x_1)\frac{p_t(z|x_0,x_1)\pi(x_0,x_1)}{p_t(z)} \\
         &=\sum_{k=1}^K\frac{1}{p_t(z)}\sum_{(x_0, x_1):\,x_1\in S_k} u_t^i(x^i,z|x_0,x_1)p_t(z|x_0,x_1)\pi(x_0,x_1|S_k)p(S_k)\\ 
         &= \sum_{k=1}^K\frac{p_t(z|S_k)p(S_k)}{p_t(z)}\sum_{(x_0, x_1):\,x_1\in S_k} u_t^i(x^i,z|x_0,x_1)\frac{p_t(z|x_0,x_1)\pi(x_0,x_1|S_k)}{p_t(z|S_k)}
     \end{align}

The inner sum is called an expert velocity, and the weighting coefficient is called a router. The equation above simply means that theoretically, the overall generating probability velocity is represented as a weighted sum of expert velocities. 


\section{Method}
Since the effectiveness of decentralized training for pure language models has already been established in previous work \citep{gururangan2023scalingexpertlanguagemodels, no-need-to-talk}, our empirical evaluation focuses on Vision Language Models as a more complex instance of autoregressive sampling. The purpose of this section is to provide a comprehensive overview of our experimental methodology. At a high level, our approach involves training individual expert models entirely independently on disjoint data subsets, and subsequently combining their predictions using a routing mechanism during the inference phase. In the following subsections, we will detail our complete decentralized pipeline: first describing the training process, which includes the data partitioning strategy, the routing mechanism, and the independent training of expert models, and finally outlining our inference strategy for utilizing these experts during generation.

\subsection{Training}
\begin{figure}[h]
    \centering
    \includegraphics[width=0.7\linewidth]{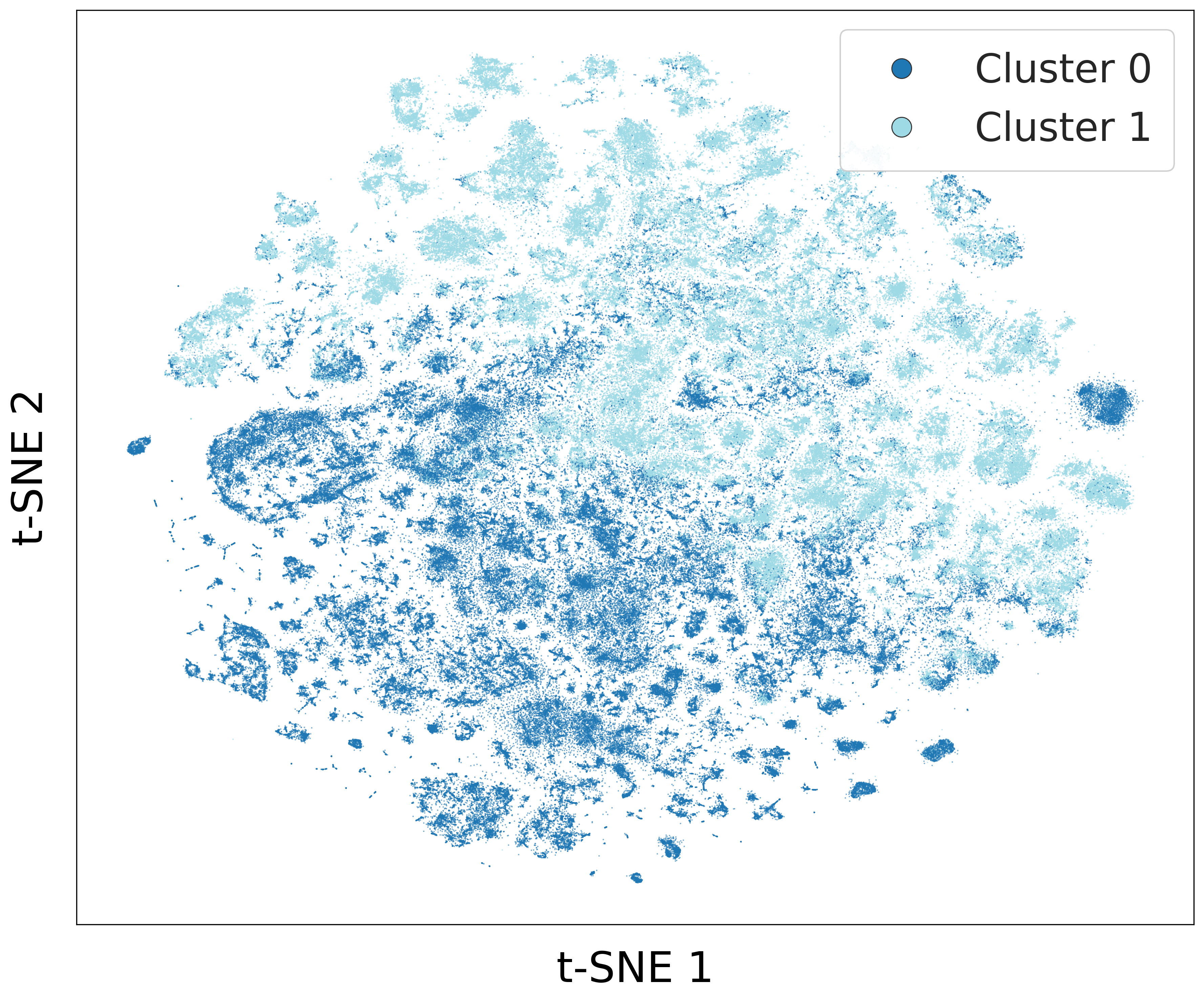}
    \caption{t-SNE 2D projection of balanced spherical $k$-means clustering with $K$=2 applied to the training samples of the InternVL models. Distinct image features from the mixed dataset are extracted using the vision encoder (CLIP-ViT-B/16) and then clustered. Each main cluster is composed of multiple smaller clusters, corresponding to groups of similar objects.}
    \label{fig:tsne-clustering}
\end{figure}

\paragraph{Data partitioning}
Following established practices in multimodal representation learning \citep{radford2021learningtransferablevisualmodels}, we use a pretrained vision encoder (CLIP) to extract image features from the unique images in the dataset. CLIP is widely adopted for this purpose because it provides robust and semantically rich visual representations. We train a spherical balanced $k$-means algorithm to cluster these features into K clusters of equal size (see Figure \ref{fig:tsne-clustering}). Specifically, we employ cosine distance between the feature vectors, as CLIP representations are naturally optimized for this metric. The centroid vectors are $L_2$-normalized to ensure they reside on a unit sphere. This normalization effectively handles the high dimensionality of CLIP features. All samples are evenly distributed among the clusters based on their distance to the centroids, ensuring each expert receives the same number of unique training images.

\paragraph{Router}
We establish the routing mechanism without introducing any additional trainable parameters or auxiliary neural networks. Instead, the K cluster centroids derived from the spherical balanced k-means algorithm serve directly as the router. For any given input sample, we extract its image features using the same pretrained CLIP vision encoder utilized during the partitioning phase. The router then computes the cosine similarity between the extracted feature vector and each fixed cluster centroid. The sample is subsequently assigned to the corresponding cluster based on these similarity scores. This approach ensures a highly efficient routing mechanism that perfectly mirrors the initial data distribution strategy.

\paragraph{Experts training}
Following the data partitioning phase, we initialize $K$ distinct expert models. Each expert is trained completely independently on its exclusively assigned data cluster. Because these experts do not require any communication or gradient synchronization with one another during the optimization process, they can be trained simultaneously across completely isolated computing nodes.

\subsection{Inference}
During inference time, for a sample with image $I$ we set cluster probabilities as:
\begin{equation}
    p_t(S_k|x) = \frac{\exp(\tau \cdot \cos(x, c_k))}{\sum_{j=1}^K \exp(\tau \cdot \cos(x, c_j))}
\end{equation}
where $x=f_\text{CLIP}(I)$ -- CLIP features vector, $\cos(\cdot, \cdot)$ is the cosine similarity, $f_\text{CLIP}$ is a feature extractor, $c_j$ are cluster centroids, $\tau$ is the temperature parameter. This implies that routing is time independent and agnostic of current token sequence state.

To balance theoretical fidelity with computational efficiency, we employ a top-$k$ strategy. The final output probabilities are top-$k$ filtered and renormalized. For the main experiments, we take $k=1$ to ensure that the ensemble inference computation budget (excluding routing cost) is the same as that of the dense model. 

\section{Experiments}

In this section, we provide a comprehensive empirical evaluation of our proposed decentralized training paradigm. We first describe the core implementation details, including the specific training dataset, the vision encoder utilized for feature extraction, and the evaluation benchmarks. Subsequently, we present the extensive experimental results for both the LLaVA and InternVL models to demonstrate the practical parity and robustness of our decentralized approach.

\subsection{LLaVA}\label{llava-experiment}

\paragraph{Training dataset}

For the finetuning of our decentralized expert models, we utilize the standard LLaVA visual instruction tuning dataset \citep{liu2024improvedbaselinesvisualinstruction}. This widely adopted corpus aggregates diverse multimodal instruction following data, including visual conversations, complex reasoning scenarios, and detailed image description tasks. By training our decentralized experts on this exact standard corpus, we ensure a rigorous foundation for evaluating our training approach.

\paragraph{Implementation details}
During the data partitioning stage, text-only samples are randomly and equally distributed between the clusters.

We train $K=2$ experts. As an initialization checkpoint for experts, we employ the LLaVA-1.5-7B model after the vision language alignment stage. We effectively perform the visual instruction tuning stage of the experts.

Each expert is trained on the same number of GPU devices as the dense baseline. However, we halve the per-device batch size to ensure that the total number of training steps remains consistent with the original model training schedule. All other hyperparameters remain unchanged.

We utilize CLIP-ViT-L/14@336px as the vision encoder for our routing mechanism. We select this specific version because the LLaVA-1.5 architecture already requires it for visual processing. Consequently, with a shared encoder implementation, the feature extraction step for routing incurs almost zero additional computational overhead during inference.

\paragraph{Evaluation benchmarks and metrics}

For our empirical evaluation, we utilize a comprehensive suite of visual reasoning and question answering benchmarks. We specifically selected these datasets because they were prominently featured in the original LLaVA paper \citep{liu2024improvedbaselinesvisualinstruction}, ensuring that our baseline comparisons are directly aligned with established literature. To assess general visual question answering, text reading, and scientific reasoning, we evaluate on VQAv2 \citep{balanced_vqa_v2}, GQA \citep{hudson2019gqanewdatasetrealworld}, VizWiz \citep{gurari2018vizwizgrandchallengeanswering}, TextVQA \citep{singh2019vqamodelsread}, and the image subset of ScienceQA \citep{lu2022learn}. For these benchmarks, performance is measured using standard evaluation accuracy, following official consensus based scoring methods. Furthermore, we measure the susceptibility of our models to object hallucination using the POPE benchmark \citep{li2023evaluatingobjecthallucinationlarge}, calculating the F1 score across its adversarial, random, and popular splits. To evaluate broader perception and cognition capabilities, we incorporate the MME benchmark \citep{fu2025mmecomprehensiveevaluationbenchmark}, reporting its standard cumulative score. Finally, for the MMBench benchmark \citep{liu2024mmbenchmultimodalmodelallaround}, we test performance on both the English and Chinese subsets, reporting the overall multiple choice accuracy.

\paragraph{Results} We evaluate our LLaVA models across a total of 8 benchmarks used by \citet{liu2024improvedbaselinesvisualinstruction}, comparing our performance directly against the official baseline statistics reported in their work. Overall, the expert ensemble achieves near-parity with the compute-matched dense baseline  (see Tables \ref{tab:llava-results-1} and \ref{tab:llava-results-2}) with small fluctuations. We observe small gains in visual-question answering (VQA) tasks and minor trade-offs in OCR tasks. The decentralized training preserves the baseline performance.

\begin{table}[h!]
    \centering
    \begin{tabular}{lccccc}
    \hline
        Method  & VQAv2 & GQA & VizWiz & SciQA-IMG & TextVQA \\
        \hline
        Dense baseline & 78.50 & \textbf{62.00} & \textbf{50.00} & 66.80 & \textbf{58.20} \\
        2 experts      & \textbf{79.99} & 61.97 & 45.53 & \textbf{67.03} & 56.67 \\
        \hline
    \end{tabular}
    \caption{LLaVA experts results on academic-task-oriented datasets}
    \label{tab:llava-results-1}
\end{table}

\begin{table}[h!]
    \centering
    \begin{tabular}{lcccccc}
    \hline
        Method & \multicolumn{3}{c}{POPE} & MME & \multicolumn{2}{c}{MMBench} \\
        & adv & rand & pop &  & en & zh \\
        \hline
        Dense baseline & 85.9 & \textbf{87.3} & 86.1 & \textbf{1510.7} & 64.3 & \textbf{58.3} \\
        2 experts      & \textbf{87.1} & 85.7 & \textbf{87.3} & 1477.38 & \textbf{65.03} & 56.87 \\
    \hline
    \end{tabular}
    \caption{LLaVA experts results on benchmarks for instruction-following LLMs}
    \label{tab:llava-results-2}
\end{table}
\subsection{InternVL}
\paragraph{Training dataset}
In the second part of our experiments, we utilize a strategically curated subset of the InternVL 2.5 \citep{chen2025expandingperformanceboundariesopensource} Stage 2 finetuning data mixture. To ensure that our models develop a comprehensive set of multimodal capabilities without requiring the entire massive dataset, we constructed this mixture by deliberately sampling one or two representative datasets from every primary task category. As shown in Table \ref{tab:internvl-dataset}, this methodology guarantees broad exposure across distinct domains. Specifically, our training mixture includes high quality data for image captioning, general visual question answering, chart comprehension, optical character recognition, knowledge based reasoning, document understanding, visual grounding, scientific reasoning, and complex conversational interactions. By intentionally selecting prominent datasets for each specific competency, we maintain a diverse training distribution that robustly evaluates our decentralized methodology across a wide spectrum of visual reasoning challenges.
\paragraph{Implementation details}

Consistent with the first experimental phase \ref{llava-experiment}, we train $K=2$ experts. We initialize the experts using the InternVL-2.5-1B-Pretrained checkpoint (post-Stage 1.5 ViT+MLP pretraining) and perform full Stage 2 finetuning. For dataset partitioning, we apply the same balanced $k$-means algorithm described previously.

To ensure a fair comparison, both the dense baseline and the experts are trained on an identical number of GPU devices with a per-device batch size of 1. We halve the number of gradient accumulation steps for the expert runs. This ensures that the total number of optimization steps remains comparable to the dense model baseline. 

Unlike the standard InternVL training pipeline, we disable dynamic resolution to accelerate training. Additionally, the number of training devices is set to 2, and the context length is capped at 8192 tokens. All other hyperparameters remain unchanged.

We utilize CLIP-ViT-B/16 as a vision encoder for dataset partitioning and routing. This results an additional 5.4\% computational overhead during inference compared to dense baseline.

We selected InternVL as an alternative to LLaVA to verify the effectiveness of our expert training strategy across different architectures and training setups. Specifically, this allows us to contrast two distinct paradigms: (1) LLaVA-1.5-7B, which uses a fixed CLIP vision encoder and only finetunes the MLP connector and LLM; and (2) InternVL 2.5-1B, which uses an Intern-ViT-300M encoder and performs full-parameter finetuning (ViT+MLP+LLM) during the instruction tuning stage.

\begin{table}[t]
    \centering
    \begin{tabular}{c|c}
    \hline
        Task type & Datasets \\
        \\ \hline\\
      Captioning   & TextCaps \citep{sidorov2020textcapsdatasetimagecaptioning}, ShareGPT4o \citep{cui2025comprehensive} \\
      \hline
      General QA & VQAv2\citep{balanced_vqa_v2}, 
      GQA\citep{hudson2019gqanewdatasetrealworld} \\
      \hline
      Chart & ChartQA\citep{masry2022chartqabenchmarkquestionanswering} \\
      \hline
      OCR & InfoVQA\citep{mathew2021infographicvqa}, TextVQA\citep{singh2019vqamodelsread} \\
      \hline
      Knowledge & KVQA\citep{shahMYP19}, A-OKVQA\citep{schwenk2022aokvqabenchmarkvisualquestion} \\
      \hline
      Document & DocVQA\citep{mathew2021docvqadatasetvqadocument} \\
      \hline
      Grounding & RefCOCO/+/g \citep{mao2016generationcomprehensionunambiguousobject, yu2016modelingcontextreferringexpressions}\\
      \hline
      Science & AI2D\citep{Kembhavi2016ADI}, ScienceQA\citep{lu2022learn} \\
      \hline
      Conversation & ALLaVA\citep{chen2024allavaharnessinggpt4vsynthesizeddata} \\
        \hline
    \end{tabular}
    \caption{InternVL-type models finetuning data mixture}
    \label{tab:internvl-dataset}

\end{table}
\paragraph{Evaluation benchmarks and metrics}
For our InternVL setting, we evaluate the models across a diverse suite of 13 benchmarks used in the original InternVL-2.5 technical report \citep{chen2025expandingperformanceboundariesopensource}, encompassing general question answering, OCR, and other multimodal tasks. We specifically selected this comprehensive array of datasets to thoroughly assess capabilities across general question answering, optical character recognition, and complex visual grounding. To assess general visual question answering and scientific reasoning, we measure standard accuracy on VQAv2 \citep{balanced_vqa_v2}, GQA \citep{hudson2019gqanewdatasetrealworld}, AI2D \citep{Kembhavi2016ADI}, and the image subset of ScienceQA \citep{lu2022learn}. For text heavy tasks requiring robust optical character recognition, we evaluate performance on DocVQA \citep{mathew2021docvqadatasetvqadocument} and InfoVQA \citep{mathew2021infographicvqa} using the Average Normalized Levenshtein Similarity to account for minor text extraction variations, whereas for TextVQA \citep{singh2019vqamodelsread} we report the standard VQA accuracy. Additionally, we measure chart comprehension on ChartQA \citep{masry2022chartqabenchmarkquestionanswering} using a relaxed accuracy metric. To quantify the susceptibility of our models to object hallucination, we report the average score across the POPE \citep{li2023evaluatingobjecthallucinationlarge} benchmark, and we capture broader perception capabilities using the cumulative score on MME \citep{fu2025mmecomprehensiveevaluationbenchmark}. Finally, to evaluate precise spatial understanding, we test performance on the RefCOCO, RefCOCO+, and RefCOCOg \citep{yu2016modelingcontextreferringexpressions} visual grounding benchmarks, reporting bounding box accuracy based on a standard Intersection over Union threshold.
\paragraph{Results} Consistent with our LLaVA findings, the decentralized expert ensemble achieves near-parity with the compute-matched dense baseline (see Tables \ref{tab:internvl-results-1}--\ref{tab:internvl-results-3}). We observe the experts yield slight gains on science question answering and diagram-based tasks, balanced by modest trade-offs on OCR tasks and broad multimodal evaluations. These results confirm that decentralized training preserves core abilities across different model architectures, with minor fluctuations reflecting the natural specialization of the experts.
\begin{table}[h!]
    \centering
    \begin{tabular}{cccccc}
    \hline
       Method  & AI2D & ChartQA & TextVQA & DocVQA & InfoVQA \\
       \hline
       Dense baseline  & 61.66 & 61.96 & \textbf{55.47} & 48.49 & \textbf{25.02} \\
       2 experts & \textbf{62.50} & \textbf{62.24} & 50.91 & 48.49 & 24.00 \\
    \hline
    \end{tabular}
    \caption{InternVL experts results on OCR, chart, and document understanding}
    \label{tab:internvl-results-1}
\end{table}

\begin{table}[h!]
    \centering
    \begin{tabular}{cccccc}
    \hline
        Method & VQAv2 & GQA & SciQA-IMG & POPE (avg) & MME \\
        \hline
        Dense baseline & \textbf{72.5} & 56.35 & 83.64 & \textbf{86.17} & \textbf{1531.7} \\
        2 experts & 72.19 & \textbf{57.36} & \textbf{83.89} & 85.55 & 1475.78\\
    \hline
    \end{tabular}
    \caption{InternVL experts results on general QA, multimodal understanding and hallucination performance}
    \label{tab:internvl-results-2}
\end{table}
\begin{table}[h!]
  \centering
  \begin{tabular}{lcccccccc}
  \hline
    & \multicolumn{3}{c}{RefCOCO} & \multicolumn{3}{c}{RefCOCO+} & \multicolumn{2}{c}{RefCOCOg} \\
    Method & val & testA & testB & val & testA & testB & val & test \\
    \hline
    Dense baseline & \textbf{77.05} & \textbf{82.98} & 71.05 & \textbf{64.82} & \textbf{73.47} & 55.61 & \textbf{72.18} & \textbf{71.34} \\
    2 experts     & 75.47 & 80.20 & \textbf{71.36}  & 62.46 & 71.29 & \textbf{56.43} &  71.94  & 70.45 \\
    \hline
  \end{tabular}
  \caption{InternVL experts results on visual grounding}
  \label{tab:internvl-results-3}
\end{table}

\subsection{Ablation Study and other results}
In this section, we conduct ablation studies under the InternVL setting over 3 factors: number of experts; the choice of vision encoder for image feature extraction for dataset partitioning and routing; and clustering algorithm. 
\paragraph{The impact of number of experts}
For the main InternVL experiments, we train $K=2$ experts. To test the impact of number of experts, we also train $K=4,6$ experts using spherical balanced k-means on CLIP ViT-B/16 features. As shown in Table \ref{tab:ablation-number-of-experts}, the $K=4$ ensemble maintains performance comparable to the dense baseline, demonstrating the robustness of the decentralized approach. The further increase in the number of experts to $K=6$ leads to only marginal changes. The slight regression compared to the $K=2$ setting is explained by data fragmentation, where the further partitioning reduces sample density per expert.
\begin{table}[h!]
    \centering
    \begin{tabular}{llccccc}
    \hline
       Method  & VQAv2 & TextVQA & MME & \multicolumn{3}{c}{RefCOCO}\\
       & & & & val & testA & testB \\
       \hline
        2 experts & \textbf{72.19} & \textbf{50.91} & \textbf{1475.78} & \textbf{75.47} & \textbf{80.20} & \textbf{71.36} \\
        4 experts & 70.96 & 47.59 & 1330.41 & 72.05 & 78.47 & 66.79 \\
        6 experts & 70.45 & 46.16 & 1353.1 & 70.92 & 76.21 & 67.30 \\
    \hline
    \end{tabular}
    \caption{The impact of number of experts}
    \label{tab:ablation-number-of-experts}
\end{table}
\paragraph{The impact of vision encoder}
In the main InternVL experiments, we use CLIP ViT-B/16 as a feature extractor for k-means input for dataset partitioning and routing. To investigate the impact of the vision encoder, we also train $K=2$ experts with balanced k-means algorithm using CLIP ViT-L/14@336px and CLIP RN50. As shown in Table \ref{tab:ablation-vision-encoder}, using a larger vision encoder leads to a marginal increase in QA (+0.4 VQAv2) and OCR (+0.81 TextVQA) performance. At the same time, using a smaller encoder leads to performance decrease on visual question answering (-4.14 VQAv2) and marginal improvement in visual grounding (+0.59 RefCOCO val).
\begin{table}[h!]
    \centering
    \begin{tabular}{llccccc}
    \hline
       Method  & VQAv2 & TextVQA & MME & \multicolumn{3}{c}{RefCOCO}\\
       & & & & val & testA & testB \\
       \hline
        ViT-B/16 & 72.19 & 50.91 & \textbf{1475.78} & 75.47 & 80.20 & \textbf{71.36} \\
        ViT-L/14@336px & \textbf{72.56} & \textbf{51.72} & 1450.04 & 74.74 & 80.75 & 71.01 \\
        RN50 & 68.05 & 47.5 & 1382.9 & \textbf{76.06} & \textbf{81.74} & 70.95 \\
    \hline
    \end{tabular}
    \caption{The impact of vision encoder}
    \label{tab:ablation-vision-encoder}
\end{table}
\paragraph{The impact of clustering algorithm}
In the main InternVL experiments, we use spherical balanced k-means clustering algorithm for dataset partitioning and routing. To test the impact of the clustering algorithm, we also train $K=2$ experts using 2-stage balanced spherical k-means algorithm inspired by \citet{mcallister2025decentralizeddiffusionmodels}. Namely, the first stage is fine clustering into $k=1024$ clusters in an unbalanced manner. The second stage is coarse clustering of fine centroids into $K=2$ coarse clusters in a balanced manner. As shown in Table \ref{tab:ablation-clustering-algorithm}, the 2-stage approach yields comparable performance with minor trade-offs on standard visual question-answering tasks. While we observe a wider performance variance on multi-task benchmarks, the overall competitive results across the majority of metrics demonstrate that the decentralized training paradigm is generally robust to the choice of clustering algorithm, though the simpler single-stage approach provides greater stability.
\begin{table}[h!]
    \centering
    \begin{tabular}{llccccc}
    \hline
       Method  & VQAv2 & TextVQA & MME & \multicolumn{3}{c}{RefCOCO}\\
       & & & & val & testA & testB \\
       \hline
        \textbf{balanced k-means} & \textbf{72.19} & 50.91 & \textbf{1475.78} & \textbf{75.47} & 80.20 & \textbf{71.36} \\
        2-stage balanced k-means & 71.82 & \textbf{51.69} & 1421.01 & 74.76 & \textbf{80.99} & 69.99 \\
        \hline
    \end{tabular}
    \caption{The impact of clustering algorithm}
    \label{tab:ablation-clustering-algorithm}

\end{table}

\section{Conclusion}

In this work, for the first time, we establish a rigorous theoretical foundation for the decentralized training of autoregressive models. We formally establish the theoretical equivalence between decentralized and centralized training objectives. To demonstrate this, we prove that the global probability generating velocity can be exactly formulated as a linear combination of independent expert velocities. To achieve this mathematical decomposition, we extend the Discrete Flow Matching framework into the discrete time domain, introducing necessary constraints such as the 1-sparse property to provide a novel alignment with autoregressive sampling. Finally, we validate our theoretical findings with extensive empirical evidence. By successfully applying our decentralized training paradigm to Multimodal Large Language Models, including the LLaVA and InternVL architectures, we demonstrate its practical effectiveness and robust performance across a diverse set of question answering benchmarks.


\bibliography{main}
\bibliographystyle{tmlr}

\appendix

\end{document}